\documentclass{article}

\usepackage[preprint]{neurips_2026}
\usepackage[utf8]{inputenc} % allow utf-8 input
\usepackage[T1]{fontenc}    % use 8-bit T1 fonts
\usepackage{hyperref}       % hyperlinks
\usepackage{url}            % simple URL typesetting
\usepackage{booktabs}       % professional-quality tables
\usepackage{amsfonts}       % blackboard math symbols
\usepackage{nicefrac}       % compact symbols for 1/2, etc.
\usepackage{microtype}      % microtypography
\usepackage{xcolor}         % colors
\usepackage{multirow}
\usepackage{graphicx}
\usepackage{subcaption}
\usepackage{amsmath}
\usepackage{tabularx}

\usepackage{dsfont} % for mathds

\title{Can't Find Waldo: Evaluating VLMs' Sensitivity\\to Image Resolution and Detail Level}

\author{%
    Alexandra Schild \\
    HPI/University of Potsdam\\
    Potsdam, Germany \\
  \texttt{aleksandra.kudaeva@hpi.de} \\
  \And
    Gerard de Melo \\
    HPI/University of Potsdam \\
    Potsdam, Germany \\
    \texttt{gerard.demelo@hpi.de}
}

\begin{document}

\maketitle

\begin{abstract}
Visual Language Models (VLMs) have achieved remarkable success across diverse tasks, yet they struggle with high-resolution inputs where critical information resides in small regions or detailed, cluttered scenes. While several approaches address this limitation, a systematic understanding of \textit{why} models fail at high resolutions is lacking. We introduce a controlled evaluation framework that disentangles resolution-related performance degradation from task difficulty through semantics-preserving transformations. We propose two simple metrics: Area Under the Scaling Curve (AUSC), which quantifies scaling robustness independent of baseline accuracy, and Prediction Variance Score (PVS), which measures resolution-induced prediction instability. Through comprehensive experiments across 5 model families and 5 benchmarks, we identify three primary failure modes: (1) information loss from downsampling at vision token limits, (2) tokenization artifacts from patch boundary shifts and positional encoding fragility under non-standard aspect ratios, and (3) attention dilution as token counts increase. Our analysis reveals that even state-of-the-art models suffer from performance drops when processing high-resolution images, with degradation patterns varying systematically by architectural family. We provide actionable insights for model architecture design and data augmentation strategies to mitigate these limitations.

%\textbf{Code:} \url{https://github.com/anonymous872/Waldo}
\end{abstract}

\section{Introduction}

Deploying Vision–Language Models (VLMs) in domains such as medical imaging~\citep{tran2025generating, dai2025pathologyvlm}, satellite imagery~\citep{luo2025largevisionlanguagemodelmeets}, and document understanding~\citep{liu2025vlm} requires reliable extraction of fine-grained visual detail across varying scales and resolutions. Yet the visual information a model can effectively access is shaped not only by image semantics, but also by how that information is encoded and processed: the input resolution, and the size, number, and position of task-relevant visual cues. As a result, a model may fail not because of an inability to interpret and reason on the input, but because critical visual evidence was lost during preprocessing, distorted by patch tokenization, or diluted across an overlong token sequence.

\begin{figure*}[h]
\centering
\includegraphics[width = 1\linewidth]{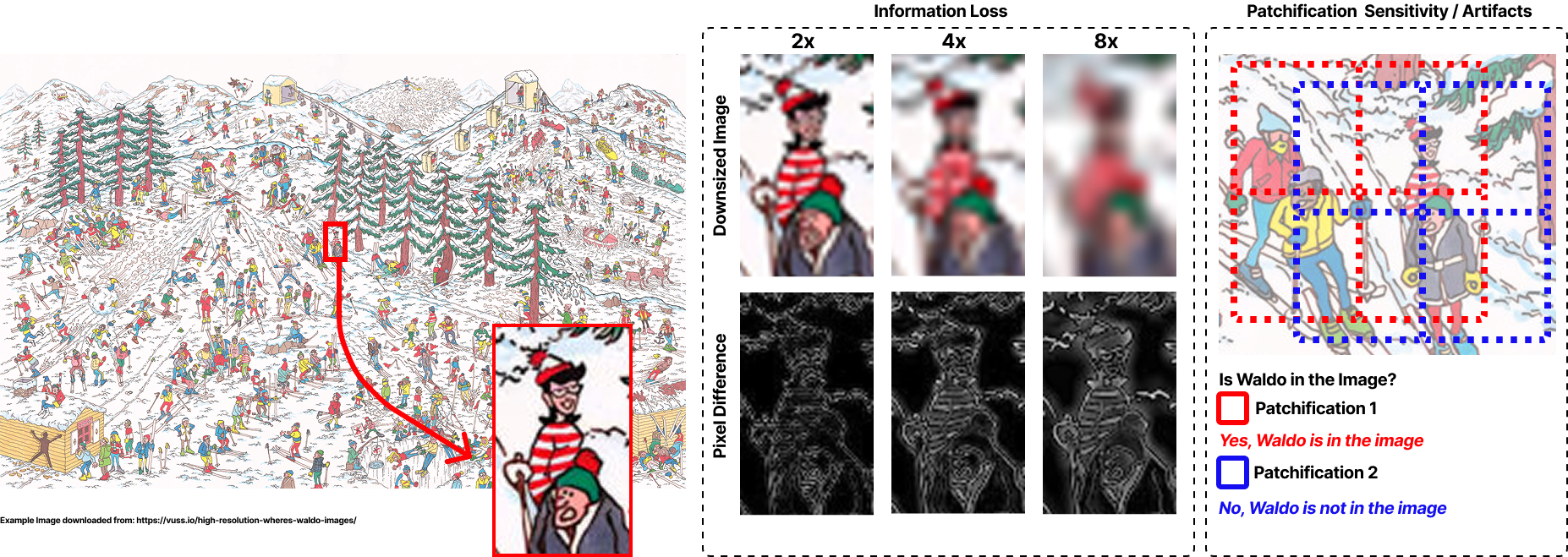}
\caption{ Information loss may occur due to various factors, including downsizing, patch artifacts, and attention dilution.}
\label{fig:information-loss}
\end{figure*}

Current VLM benchmarks span heterogeneous image collections with varying resolutions, aspect ratios, and information densities ~\citep{fu2024mmecomprehensiveevaluationbenchmark, zhang2024mathversedoesmultimodalllm, yu2024mmvetevaluatinglargemultimodal}. Although intended to reflect real-world conditions, this diversity introduces a confound: performance differences may reflect how well each architecture handles incidental image properties rather than differences in perceptual or reasoning ability, complicating fair comparison and failure attribution. To isolate these effects, we introduce a diagnostic framework that applies controlled, semantics-preserving transformations to existing VQA benchmarks, independently varying image resolution and aspect ratio, level of detail, and the spatial position of task-relevant content. The controlled combination of these transformations further enables us to disentangle different sources of performance degradation and identify which failure modes individual model architectures are particularly susceptible to.

On top of this framework, we propose two complementary metrics. \textit{Area Under the Scaling Curve} (AUSC) quantifies how model performance degrades as resolution increases, and \textit{Prediction Variance Score} (PVS) measures instability in model predictions across resolution variants of the same image, revealing that the model possesses the underlying reasoning capacity but that its answer is contingent on tokenization artifacts.

Through extensive experiments across five architecturally distinct model families and multiple model sizes, we identify and characterize three primary failure modes:

\begin{enumerate}
    \item \textbf{Downsampling-induced information loss}, where fine-grained detail is irreversibly discarded when image resolution exceeds the model's vision token budget.
    \item \textbf{Tokenization artifacts} arising from patch boundary misalignment and information density mismatch, causing non-monotonic, resolution-dependent prediction instability.
    \item \textbf{Attention dilution}, where increasing token counts cause the LLM decoder to lose effective focus on spatially localized task-relevant cues.
\end{enumerate}

The onset and severity of each failure mode varies by architecture, reflecting differences in dynamic resolution strategies, token budget management, and attention mechanisms—differences which standard single-resolution benchmarks are blind to.

Our contributions are threefold:

\begin{itemize}
    \item A controlled evaluation framework isolating visual encoding sensitivity from reasoning ability via semantics-preserving variations in resolution, aspect ratio, and spatial position of task-relevant cues.
    \item Two architecture-agnostic metrics (AUSC and PVS) quantifying resolution robustness independently of baseline accuracy. 
    \item A systematic analysis framework leveraging cross-scenario prediction trajectories to identify patterns of instability and provide insights into architectural and training factors potentially contributing to resolution sensitivity, together with possible mitigation strategies.
\end{itemize}

\section{Related Work}

\subsection{Scaling Vision Encoders to High Resolution}

Processing high-resolution images in Vision–Language Models involves a fundamental tension between preserving fine-grained visual detail and managing computational cost~\citep{bakhtiarnia2024efficient}. Early models, including LLaVA~1.5 and BLIP-2, resample inputs to a fixed resolution, discarding information before reasoning~\citep{liu2023improvedllava,li2023blip2bootstrappinglanguageimagepretraining}. Dynamic-tiling methods instead process local crops alongside a global thumbnail~\citep{liu2024llavanext,xu2024llavauhdlmmperceivingaspect,li2024monkeyimageresolutiontext,grattafiori2024llama3herdmodels}, improving flexibility but increasing computation and retaining implicit resolution and aspect-ratio constraints. Native-resolution models remove fixed input dimensions at the cost of variable token counts~\citep{agrawal2024pixtral12b,wang2024qwen2vlenhancingvisionlanguagemodels,bai2025qwen25vltechnicalreport}. Recent content-adaptive tokenization and selective-encoding approaches further reduce computation by allocating tokens according to visual complexity or task relevance~\citep{shen2025catcontentadaptiveimagetokenization,zhou2025dynrslvlmenhancingautonomousdriving,shi2025scalingvisionpretraining4k}, but may introduce new sensitivities to image layout and content position.

\subsection{Evaluating High-Resolution Visual Processing}

Recent benchmarks expose persistent limitations in fine-grained visual processing. MME-RealWorld reports substantial degradation on high-resolution images~\citep{zhang2024mme}, while HRScene identifies spatially systematic failures that intensify with image size~\citep{zhang2025hrscene}. VisualOverload evaluates dense scenes and reveals severe failures in counting, OCR, and logical consistency~\citep{gavrikov2025visualoverload}. VER-Bench tests evidence extraction from small visual regions and distinguishes evidence coverage from reasoning quality~\citep{qiang2025ver}. V*Bench and Visual Haystacks similarly emphasize precise spatial grounding and retrieval over raw model scale~\citep{wu2024v,wu2024visual}. However, these benchmarks generally evaluate each image at a single resolution, conflating resolution-induced failures with semantic complexity, task difficulty, and image content.

\subsection{Evaluating Robustness in Vision--Language Models}

Robustness studies further connect VLM failures to architectural design. VLM-RobustBench finds that geometric and resampling perturbations are more damaging than perceptually severe photometric corruptions~\citep{saxena2026vlm}. V$^2$R-Bench attributes sensitivity to scale and spatial variation to error accumulation in the vision encoder and insufficient multimodal alignment~\citep{fan-etal-2025-unveiling}, while HRScene links resolution-dependent spatial failures to uneven token compression under dynamic patching~\citep{zhang2025hrscene}. These works identify important failure patterns but do not systematically isolate resolution, detail, and spatial sensitivity or attribute their effects to individual architectural components.

\section{Background}
\subsection{Problem Formulation}

Let $\mathcal{B} = \{(x_i, y_i)\}_{i=1}^N$ denote a VQA benchmark, where $x_i \in \mathbb{R}^{H_i \times W_i \times 3}$ is an image and $y_i$ is the corresponding question-answer pair. A VLM $f$ maps each pair to a predicted answer $\hat{z}_i = f(x_i, y_i)$, evaluated against ground truth $z_i$. Standard benchmarks measure aggregate accuracy $\frac{1}{N}\sum_i \mathbf{1}[\hat{z}_i = z_i]$ at a single fixed resolution per image, implicitly assuming performance is invariant to image scale and tokenization structure. We challenge this assumption and study model behavior under a family of semantics-preserving transformations $\mathcal{T} = \{T_1, \ldots, T_M\}$, where each $T_j$ modifies an image along a controlled axis (pixel density or token allocation) without altering the visual content relevant to $y_i$.

Processing a visual query involves a sequential causal chain:

\begin{equation}
x \xrightarrow{\text{preproc.}} \text{pixels}
\xrightarrow{\text{tokenization}}
\substack{\text{vision}\\\text{tokens}}
\xrightarrow{\text{encoder}} \text{visual features}
\xrightarrow{\text{projection}}
\substack{\text{aligned}\\\text{tokens}}
\xrightarrow{\text{LLM}} \hat{z}
\label{eq:causal_chain}
\end{equation}

Evaluating performance across controlled transformations and scales provides a structured view of model behavior beyond aggregate accuracy. It enables us to decompose performance into more informative components, identify likely failure modes and attribute them to specific model components.

\subsection{Failure Modes}

\begin{figure}[t]
\centering
\includegraphics[width=1\linewidth]{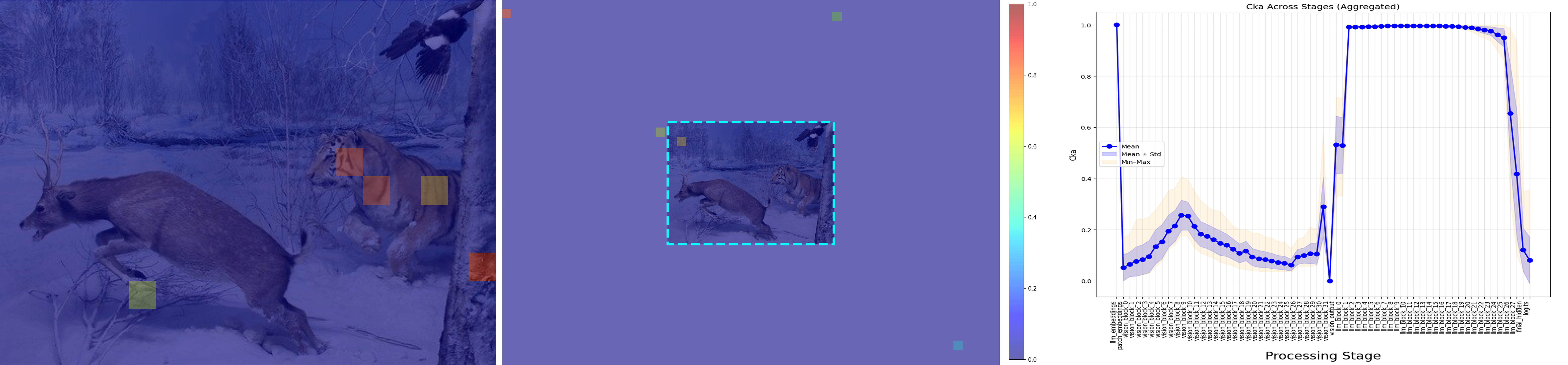}
\caption{\textbf{On the left:} Illustration of attention dilution in Qwen2-VL-2B. Despite
identical visual content within the cyan bounding box, attention weights diffuse across uninformative
blank regions, reducing concentration on task-relevant content and degrading performance; 
\textbf{On the right:} Centered Kernel Alignment (CKA) calculated for all layers of the Qwen2-VL-2B VLM comparing internal representations for two images: (1) original image padded with white field, (2) upscaled image}
\label{fig:failure_modes}
\end{figure}

We identify three prominent sources of performance degradation across VLM components: (1) information loss induced by image downsizing during preprocessing, (2) tokenization artifacts introduced by the visual encoder, and (3) attention dilution in the LLM. 

\subsubsection{Downsizing-Induced Information Loss}

Downsizing-induced information loss arises when the preprocessing pipeline reduces image resolution before visual encoding. This is a major source of performance degradation in models that apply aggressive resizing (e.g., LLaVA-1.5) and also in models that otherwise support high-resolution inputs once their vision token budget is exceeded (e.g. Pixtral, Qwen-family). In the latter case, the effective input is still constrained through automatic downsampling or cropping, which limits the spatial detail available to downstream components.

\subsection{Tokenization Artifacts}

Vision Transformers partition images into fixed-size patches (typically 14×14 or 16×16 pixels) that serve as input tokens. This discrete tessellation introduces a fundamental dependency: Model performance can vary based on where patch boundaries fall relative to semantic content, independent of the information itself. We identify two distinct sources of tokenization artifacts: (a) \textbf{spatial shift artifacts}, where translating the patch grid by a few pixels changes which visual features are split across patch boundaries, and (b) \textbf{compression artifacts}, where downsampling increases the semantic density per patch, forcing individual tokens to encode different level of detail.

To examine how these artifacts propagate through the model, we compared layer-wise representations of the same visual content placed at different background locations in Qwen2-VL-2B-Instruct. The resulting representations become more similar in later layers of the vision encoder, but positional differences remain clearly encoded. In the LLM, visual information has little influence in the early and middle layers, where representations are largely dominated by text; location-dependent differences emerge only in the later decoder layers (\autoref{fig:failure_modes}).

\subsection{Attention Dilution}

Native resolution models, such as the Qwen-VL series, process images without resizing, converting each image into a variable number of visual tokens proportional to its dimensions. As padding increases the image size, the resulting token sequence grows quadratically, forcing the model's attention mechanism to distribute its capacity across increasingly sparse representations of the same visual content. This attention dilution occurs when the attention distribution becomes diffuse across many redundant tokens% representing blank canvas
, reducing the relative weight assigned to semantically meaningful regions. To illustrate this effect, we analyzed attention maps for images where the correct answer flipped to an incorrect one after introducing image transformations (Figure \ref{fig:failure_modes}).

%-------------------------------------------------------------------------
\section{Evaluation Framework}
%-------------------------------------------------------------------------

We design two complementary transformation protocols, each targeting a distinct stage of the processing pipeline. Together, they form a \textit{decomposition along the causal chain}: Pixel-Resolution probes the full pipeline response to increased pixel density; Token Resolution isolates tokenization-level effects by holding inter-pixel content fixed. The difference in model response between the scenarios is therefore causally informative.

\begin{figure}[t]
\centering
\includegraphics[width=1\linewidth]{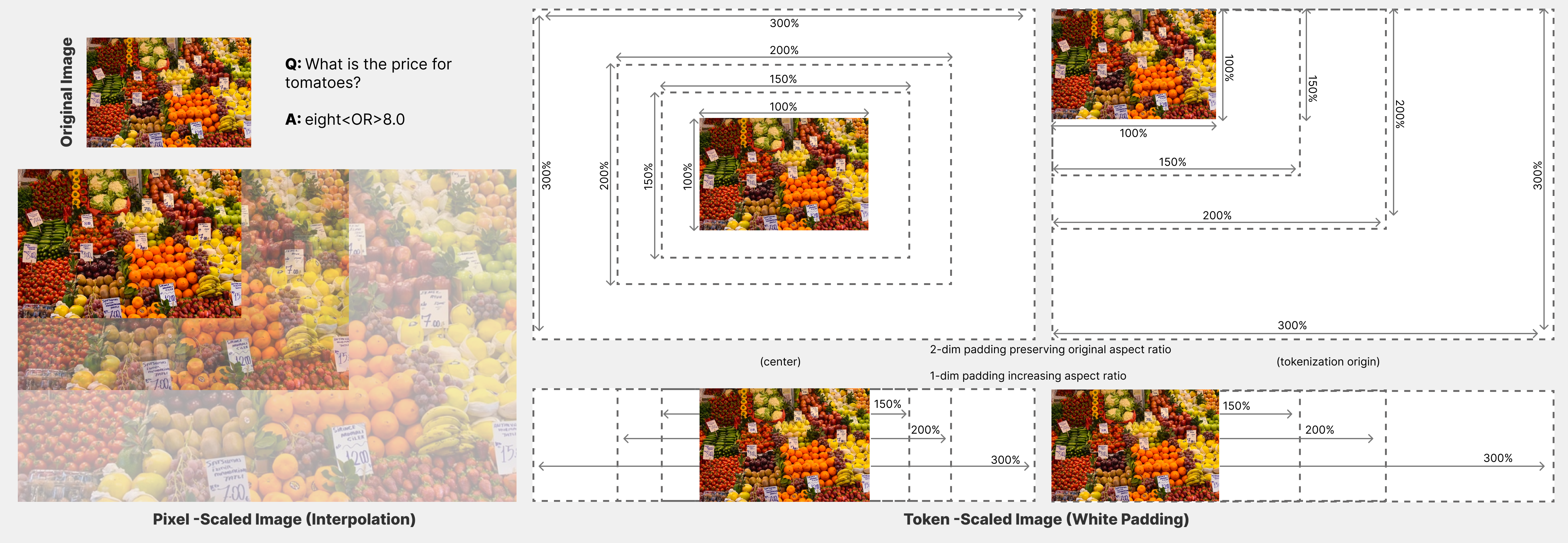}
\caption{Pixel- and Token- Resolution Scaling Example}
\label{fig:method}
\end{figure}

\subsection{Pixel-Level Resolution Scaling}

We generate higher-resolution variants of each benchmark image via bicubic interpolation upsampling operator. This transformation scales spatial dimensions uniformly by factor $s$, preserving global semantic content and the relative scale of all objects within the frame. No additional task-relevant information is introduced: objects subtend the same proportion of the image regardless of $s$.

From the model's perspective, interpolation reduces \textit{information density per token}: the same semantic content is distributed across more pixels and therefore more vision tokens, decreasing the task-relevant signal per token. Pixel resolution, therefore, probes how the model perceives changes in information density, while detail level stays constant.

\subsection{Token-Level Resolution Scaling via Canvas Padding}

We increase effective image resolution by embedding the original image within a larger white canvas, leaving pixel content entirely unchanged. This forces the model's preprocessing pipeline to handle a larger image, typically triggering more vision tokens, shifted patch boundaries, and—in dynamic tiling models—a different tile arrangement, all without any change to the inter-pixel distances in the relevant image region. 
This transformation isolates tokenization dynamics—patch boundary misalignment, increased token sequences, and attention dilution—from any pixel-level change, enabling attribution of performance changes to the tokenization stage and beyond.

It is important to note that, while canvas padding may appear out-of-distribution, we empirically confirm that across all models it is not adversarial (Appedix~\ref{sec:ablation}). 

\paragraph{Spatial position variants.}
To further probe the effect of patch boundary alignment and spatial attention bias, we place the original image at two positions on the canvas: \textit{top-left}, and \textit{center}. These positions were selected to expose distinct mechanistic effects with minimal computational overhead. Top-left preserves the original patch boundary alignment for models that do not resize, serving as a controlled baseline. Center placement tests spatial attention bias and tile-boundary splitting.

%-------------------------------------------------------------------------
\subsection{Metrics}
%-------------------------------------------------------------------------

We propose two complementary metrics to quantify resolution scaling behavior, designed to be architecture-agnostic and independent of baseline performance level. Both metrics condition on subsets of $\mathcal{B}$ stratified by correctness at original resolution, and we recommend $|\mathcal{B}_{\text{correct}}| \geq 75$ and $|\mathcal{B}_{\text{wrong}}| \geq 75$ for statistical reliability.

\subsubsection{Area Under the Scaling Curve (AUSC)}

\paragraph{Motivation.}
Raw accuracy conflates baseline capability with scaling robustness. 
AUSC isolates scaling behavior from baseline performance by conditioning exclusively on examples the model has proven to have sufficient reasoning ability to answer.

\paragraph{Definition.}
Let $\mathcal{B}_{\text{correct}} \subseteq \mathcal{B}$ denote the subset of examples answered correctly at original resolution:
\begin{equation}
    \mathcal{B}_{\text{correct}} = \{(x_i, y_i) \in \mathcal{B} \mid f(x_i, y_i) = z_i \}
    \label{eq:correct_batch}
\end{equation}
For transformation sequence $\mathcal{T} = \{T_1, \ldots, T_M\}$, let $\mathrm{Acc}(T_j)$ denote accuracy on $\{(T_j(x_i), y_i) \mid (x_i,y_i) \in \mathcal{B}_{\text{correct}}\}$. AUSC is computed via the trapezoidal rule:
\begin{equation}
    \mathrm{AUSC} = \frac{1}{M-1} \sum_{j=1}^{M-1} \frac{1}{2}\Bigl[\mathrm{Acc}(T_j) + \mathrm{Acc}(T_{j+1})\Bigr]
    \label{eq:AUSC}
\end{equation}

%\paragraph{Interpretation.}
AUSC $\in [0,1]$, where 1 indicates perfect robustness across all transformations and 0 indicates complete brittleness.

\subsubsection{Prediction Variance Score (PVS)}
\label{sec:pvs}

\paragraph{Motivation.}
A complementary signal is whether resolution changes can \textit{rescue} initially incorrect predictions, indicating that the model possesses the underlying reasoning capacity, but that its answer is sensitive to tokenization or encoding artifacts.

\paragraph{Definition.}
Let $\mathcal{B}_{\text{wrong}} = \mathcal{B} \setminus \mathcal{B}_{\text{correct}}$. For each transformation $T_j$:
\begin{equation}
    \mathrm{PVS}(T_j) = \frac{1}{|\mathcal{B}_{\text{wrong}}|} \sum_{(x_i,y_i) \in \mathcal{B}_{\text{wrong}}} \mathbf{1}\bigl[f(T_j(x_i),\, y_i) = z_i\bigr]
    \label{eq:PVS}
\end{equation}

%\paragraph{Interpretation.}
PVS $\in [0,1]$ measures the fraction of initially incorrect predictions corrected by $T_j$. Values near 0 indicate that errors are stable and likely reflect genuine reasoning limitations. Higher values
reveal resolution-dependent instability.
%-------------------------------------------------------------------------
\subsection{Attribution Analysis}
%-------------------------------------------------------------------------

The two transformation protocols jointly enable attribution of observed sensitivity to specific components of the processing pipeline. We track each example's answer correctness as a binary sequence across scaling levels. Based on that, we identify diagnostically distinct trajectory patterns, summarized in Table~\ref{tab:patterns}, with cross-scenario fingerprints enabling attribution.

\begin{table}[h]
\centering
\small
\caption{Answer trajectory patterns, their diagnostic interpretation, and the cross-scenario fingerprint enabling attribution.}
\label{tab:patterns}
\setlength{\tabcolsep}{4pt}
\begin{tabularx}{\columnwidth}{lllX}
\toprule
\textbf{Pattern} & \textbf{Trajectory} & \textbf{Attribution} & \textbf{Fingerprint} \\
\midrule
A & $1\cdots1\,0\cdots0$ (threshold) & Downsampling loss & Pixel Resolution scaling only, at token budget \\
B & Alternating & Boundary / density & Token Resolution Scaling $\to$ boundary; Pixel Resolution Scaling $\to$ density \\
C & $0\cdots0\,1\cdots$ & Tokenization masking & PVS signal; both scenarios \\
D & $1\,1\,0\,0\cdots$ & Attention dilution & Both scenarios, same token count \\
E & $0\cdots0$ & Reasoning failure & Excluded from metrics \\
F & $1\cdots1$ & Robust correct & AUSC numerator \\
G & Position-varying & Spatial bias / tiling & Token-Resolution Scaling, position-stratified \\
\bottomrule
\end{tabularx}
\end{table}

A more detailed version of the analysis complemented with VLM architecture attribution and recommended measures is provided in Appendix~\ref{future_work}.

%-------------------------------------------------------------------------
\section{Experiments}
%-------------------------------------------------------------------------

\subsection{Experimental Set-up}

\textbf{Scaling Range.} We empirically select an image-scale range that provides enough granularity to expose encoder instabilities while keeping the computational cost manageable. We evaluate models over a fixed range of image scales (1.5×–6×) to measure how performance changes with increasing visual detail. The lower end is sufficient to reveal sensitivity to small changes in resolution, while the upper end approaches the practical limits of current VLMs, where larger inputs become increasingly constrained by context length and memory. This range provides enough coverage to capture meaningful scaling behavior without making evaluation prohibitively expensive.

\textbf{Models.} We chose several commonly used open and closed source models, specifically LLaVa-1.5 \citep{liu2024llavanext}, Llama-3.2 \citep{grattafiori2024llama3herdmodels}, Qwen-2-XB \citep{wang2024qwen2vlenhancingvisionlanguagemodels}, Qwen-2.5-XB \citep{Qwen2.5-Omni}, Qwen-3-XB \citep{yang2025qwen3technicalreport}, and Pixtral \citep{agrawal2024pixtral12b}.%, and GPT-5 nano \citep{openai2024gpt4o}. 

\textbf{Benchmarks.} To assess differences in scaling across different types of VQA, we chose the MMVet \citep{yu2024mmvetevaluatinglargemultimodal}, CharXiv \citep{wang2024charxivchartinggapsrealistic}, MME \citep{fu2024mmecomprehensiveevaluationbenchmark}, MMStar \citep{chen2024rightwayevaluatinglarge}, and MathVerse \citep{zhang2024mathversedoesmultimodalllm} benchmarks for further analysis. Our choice is motivated by the desire to create a diverse set of image-question pairs across different aspects of model performance and image sizes. However, the proposed method is benchmark-agnostic and could be applied to any set of benchmarks that best matches the evaluation purpose. Further per-benchmark analysis is provided in Appendix \ref{sec:per_benchmark_analysis}.

\textbf{Sampling Strategy.} Similarly to \citet{lee2024vhelm}, we randomly sample 200 instances from each of the benchmarks in order to alleviate financial and temporal constraints. The overall size of the evaluation sample used in the experiments is 1,000 image-question pairs.

\textbf{Evaluation Protocol} To evaluate correctness of VQA answers we use a hybrid approach combining answer parsing for cases where it is possible and LLM-as-a-judge (gpt-5-nano) for more complex cases. Detailed evaluation details are provided in Appendix \ref{app:llm_judge}.

\subsection{Experimental Results}

\subsubsection{Accuracy Analysis}

\begin{figure}[h]
\centering
\includegraphics[width=\linewidth]{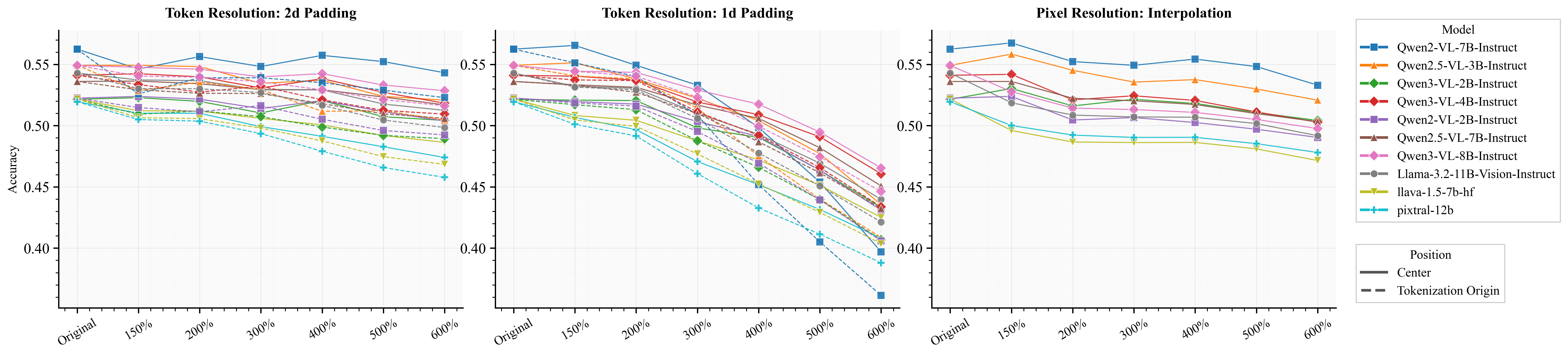}   
\caption{Accuracy degradation under different scaling factors.}
\label{fig:accuracy}
\end{figure}

We report accuracy across five transformation conditions designed to isolate distinct failure modes (\autoref{fig:accuracy}). Under 2D padding, degradation is modest ($\leq$10\%), with native-resolution models barely affected, confirming that preprocessing downsampling, not the added padding itself, drives information loss. Placing the image at the top-left corner (the tokenization origin) consistently underperforms centering, revealing a learned spatial bias toward centered content rather than position-invariant perception, in contrast to the findings by \citet{zhang2025hrscene}.

The most revealing contrast is between 1D and 2D padding. Despite producing fewer total tokens, 1D padding causes more severe degradation (5–18\%). This directly disentangles two candidate failure modes: if attention dilution over uninformative tokens were dominant, 2D padding should perform worse. Instead, the extreme aspect ratios push positional encodings out of distribution, identifying positional encoding fragility as an independent failure mode. 

Bicubic interpolation reveals a non-monotonic accuracy profile. Several models improve at 1.5× before declining, while others degrade immediately, suggesting model-specific optimal information density ranges shaped by pre-training distributions. We further investigate this by analyzing AUSC, PVS and answer trajectories.

\begin{table*}[t]
\centering
\caption{Orig. Acc., AUSC, and PVS across augmentation settings for different VLMs.}
\label{tab:vlm_ausc_pvs}
\setlength{\tabcolsep}{3pt}
\scriptsize
\resizebox{\textwidth}{!}{%
\begin{tabular}{l c ccccc ccccc}
\toprule
& & \multicolumn{5}{c}{AUSC} & \multicolumn{5}{c}{PVS} \\
\cmidrule(lr){3-7} \cmidrule(lr){8-12}
\textbf{Model} & \textbf{Orig. Acc.}
& \textbf{1D-C} & \textbf{1D-T} & \textbf{2D-C} & \textbf{2D-T} & \textbf{Intra}
& \textbf{1D-C} & \textbf{1D-T} & \textbf{2D-C} & \textbf{2D-T} & \textbf{Intra} \\
\midrule
Qwen2-VL-2B         & 0.448 & 0.752 & 0.716 & 0.845 & 0.826 & 0.827 & 0.368 & 0.368 & 0.290 & 0.307 & 0.300 \\
Qwen2-VL-7B         & 0.562 & 0.786 & 0.733 & 0.850 & 0.830 & 0.851 & 0.385 & 0.397 & 0.353 & 0.348 & 0.334 \\
\midrule
Qwen2.5-VL-3B       & 0.536 & 0.830 & 0.814 & 0.832 & 0.823 & 0.846 & 0.368 & 0.352 & 0.361 & 0.319 & 0.352 \\
Qwen2.5-VL-7B       & 0.604 & 0.858 & 0.853 & 0.870 & 0.864 & 0.871 & 0.382 & 0.397 & 0.359 & 0.351 & 0.349 \\
\midrule
Qwen3-VL-2B         & 0.466 & 0.803 & 0.799 & 0.827 & 0.817 & 0.836 & 0.414 & 0.376 & 0.416 & 0.394 & 0.445 \\
Qwen3-VL-4B         & 0.600 & 0.836 & 0.820 & 0.855 & 0.851 & 0.775 & 0.454 & 0.475 & 0.475 & 0.454 & 0.429 \\
Qwen3-VL-8B         & 0.627 & 0.842 & 0.832 & 0.866 & 0.858 & 0.683 & 0.428 & 0.414 & 0.414 & 0.384 & 0.398 \\
\midrule
Llama-3.2-11B-V & 0.499 & 0.569 & 0.578 & 0.719 & 0.704 & 0.777 & 0.493 & 0.483 & 0.477 & 0.456 & 0.475 \\
\midrule
Llava-1.5-7b-hf              & 0.352 & 0.723 & 0.589 & 0.673 & 0.593 & 0.751 & 0.307 & 0.317 & 0.365 & 0.334 & 0.306 \\
\midrule
Pixtral-12b                  & 0.499 & 0.521 & 0.507 & 0.707 & 0.694 & 0.890 & 0.432 & 0.483 & 0.422 & 0.450 & 0.349 \\
\bottomrule
\end{tabular}%
}
\end{table*}

\subsubsection{AUSC and PVS}

While AUSC measures how well models maintain correct predictions across resolutions, PVS reveals cases where resolution changes rescue initially incorrect predictions. High PVS values indicate that a model's failures are partially attributable to resolution-dependent artifacts rather than fundamental reasoning limitations. Table \ref{tab:vlm_ausc_pvs} presents AUSC and PVS metrics across different scaling factors for representative models from each architectural family. We observe three distinct patterns:

\textbf{Positional fragility dominates token dilution}. The AUSC gap between 1D and 2D padding is consistent and substantial—up to 20 points for Pixtral (0.521 vs.\ 0.707) and Llama (0.569 vs.\ 0.719). Since 1D padding produces fewer total tokens yet causes greater degradation, aspect ratio distortion of positional encodings is demonstrably more damaging than attention dilution over uninformative padding regions. Generational improvements within the Qwen family (Qwen2: 0.752 → Qwen2.5: 0.830) suggest this fragility is partially addressable through diverse aspect ratio training.

\textbf{Scale robustness and positional robustness are independent axes.} Pixtral achieves the highest interpolation AUSC (0.890) yet the lowest 1D padding AUSC (0.507)—highly resilient to density changes but fragile to positional shift. Qwen3-VL-8B exhibits the inverse: strong padding tolerance but degraded interpolation stability (0.683), indicating possible overfitting to a narrow information density range. This dissociation confirms that these represent fundamentally different failure mechanisms requiring distinct solutions.

\textbf{PVS combined with AUSC reveals prediction volatility.} Models with simultaneously high PVS and low AUSC—such as Llama-3.2 and Pixtral under 1D padding—exhibit volatile representations where predictions flip in both directions, suggesting near-decision-boundary uncertainty. In contrast, LLaVA-1.5 shows low PVS with moderate AUSC, indicating rigid confidence in both correct and incorrect predictions. Notably, PVS values of 0.30–0.49 across all conditions imply that a substantial fraction of baseline errors stem from suboptimal default tokenization rather than fundamental capability gaps.

\section{Conclusion}

VLM evaluation on VQA tasks is typically based solely on accuracy, implicitly treating all errors as failures of multimodal reasoning. Our analysis shows that this view is incomplete: only a small fraction of incorrect predictions stem from fundamental reasoning limitations. Many errors arise from information loss introduced during image preprocessing, instabilities in the vision encoder, and attention dilution caused by longer visual token sequences. These findings suggest that improving VLM performance requires not only stronger reasoning modules, but also more robust visual encoders, adapters, and evaluation protocols that disentangle perception, tokenization, and reasoning effects.

%\begin{ack}
%Acknoledgements
%\end{ack}

%%%%%%%%%%%%%%%%%%%%%%%%%%%%%%%%%%%%%%%%%%%%%%%%%%%%%%%%%%%%
\newpage
\small
\bibliographystyle{apalike}
\bibliography{references}

%%%%%%%%%%%%%%%%%%%%%%%%%%%%%%%%%%%%%%%%%%%%%%%%%%%%%%%%%%%%
\newpage
\appendix

\section{Limitations}\label{appendix:limitations}

We caution that the newly introduced metrics are dependent on the chosen image–question pairs and hence can solely be used for comparing between models evaluated on the same sample and for determining reasons for potential model performance instabilities, not as an absolute numeric measure.

The identified failure modes are not mutually exclusive, as they per design identify the first component along the VLM processing chain where the crucial information is likely lost.

\section{Future Work Discussion}\label{future_work}

\begin{table*}[h]
\centering
\small
\caption{Failure mode attribution to pipeline components and recommended interventions. Checkmark weight indicates strength of component involvement.}
\label{tab:attribution}
\setlength{\tabcolsep}{4pt}
\begin{tabular}{p{2cm}cccp{2.6cm}p{2.6cm}p{2.6cm}}
\toprule
\textbf{Failure Mode} & \textbf{Img.} & \textbf{VL} & \textbf{LLM} & \textbf{Architecture} & \textbf{Training Tasks} & \textbf{Training Data} \\
\textbf{(Pattern)} & \textbf{Enc.} & \textbf{Align.} & & \textbf{Changes} & \textbf{Changes} & \textbf{Changes} \\
\midrule
\textbf{1.\ Down- sampling Loss} 
    & $\checkmark\checkmark\checkmark$ & & 
    & Adaptive resolution; hierarchical/pyramid encoder; multi-scale feature extraction; increase native resolution (1024$\times$1024+)
    & Pre-train on high-res tasks; multi-scale prediction tasks
    & High-resolution datasets; preserve original resolutions \\

\textbf{2a.\ Patch Boundary Sensitivity}
    & $\checkmark\checkmark\checkmark$ & &
    & Overlapping patches; continuous tokenization (Perceiver); sliding window approaches
    & Tasks requiring cross-patch reasoning; random patch offset augmentation
    & Annotations spanning patch boundaries \\

\textbf{2b.\ Information Density Mismatch}
    & $\checkmark\checkmark$ & $\checkmark$ & $\checkmark\checkmark$
    & Adaptive token allocation; variable patch sizes; learned token pruning/expansion
    & Detail-focused pretraining; dense region description; small object detection
    & Variable-density images; sparse/dense scene mixtures; fine-grained datasets \\

\textbf{3.\ Attention Dilution}
    & & & $\checkmark\checkmark\checkmark$
    & Hierarchical attention; sparse/local attention; region-of-interest modules; cross-attention gating
    & Spatial attention supervision; localization-aware tasks; progressive detail focusing
    & Ground-truth attention maps; saliency-annotated data \\

\textbf{4.\ Vision–Language Alignment Gap}
    & $\checkmark$ & $\checkmark\checkmark\checkmark$ & $\checkmark$
    & Richer projection layers; multi-layer alignment; feature-preserving projectors
    & Fine-grained VQA; contrastive attribute learning; pixel-word grounding; referring expression tasks
    & Detailed attribute annotations; fine-grained VQA; grounding datasets (RefCOCO); descriptive region captions \\

\textbf{5.\ Spatial Attention Bias}
    & $\checkmark$ & $\checkmark$ & $\checkmark\checkmark$
    & Position-aware attention; learned spatial priors; tile-aware cross-attention
    & Position-randomized training; spatial localization tasks
    & Spatially balanced datasets; annotations at varied image positions \\
\bottomrule
\end{tabular}
\end{table*}

\section{Additional Analysis}

\subsection{Is Padding an Out-of-Distribution Artifact?}
\label{sec:ablation}

Since padding augmentation introduces images with large uniform borders, we asked whether the observed degradation is partly driven by the Out-Of-Distribution nature of the augmented images than by reduced effective information density.

To test this, we compare model performance on original high-resolution images~\citep{wu2024v} against versions where all irrelevant content is masked out and replaced with a uniform white background, while preserving the question-relevant region. If white padding were inherently harmful, performance should drop in the masked setting. Instead, all models perform better when irrelevant content is replaced with white space.

This shows that white regions are not intrinsically OOD for VLMs and are not the primary cause of the degradation observed under white-padding augmentation. The performance drop is instead driven by reduced effective information density.

\begin{figure}[h]
\centering
\includegraphics[width=1\linewidth]{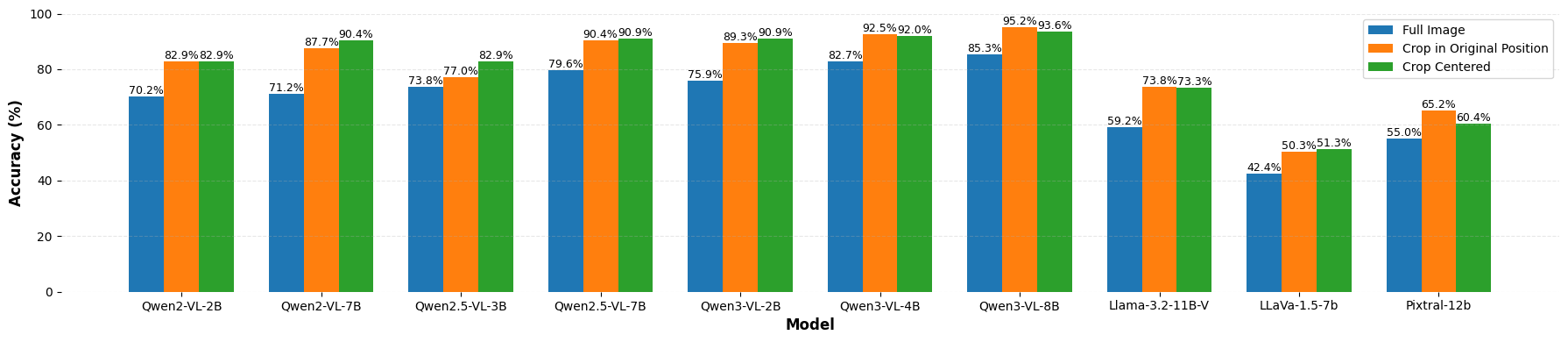}
\caption{All models perform better when irrelevant content is replaced with white background, showing that white padding itself is not the primary cause of performance degradation.}
\label{fig:ood_ablation}
\end{figure}

\subsection{Limitations of Single-Resolution VQA Accuracy}

Standard VQA accuracy evaluates whether a model answers each question correctly for a single image representation. To examine what this aggregate score obscures, we conduct an additional analysis of prediction outcomes across semantically preserving image transformations. The results reveal that accuracy alone cannot distinguish persistent model failures from predictions that change because of incidental image statistics. This distinction is particularly important for high-resolution inputs, for which detail size, resizing, and patch alignment can substantially affect the visual evidence available to the model.

\begin{figure}
    \centering
    \includegraphics[width=0.9\linewidth]{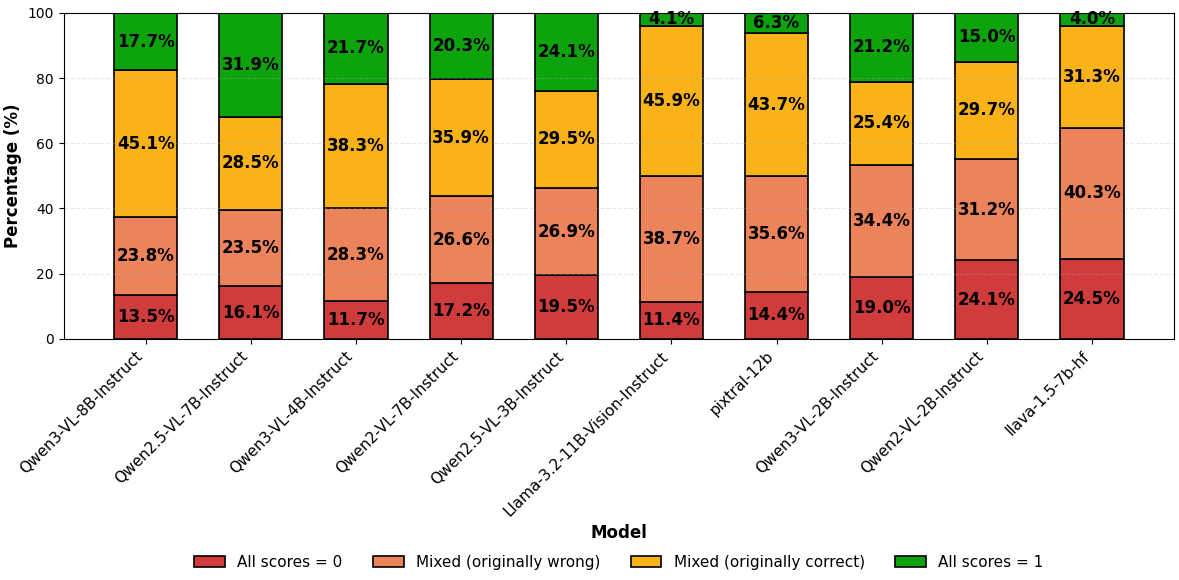}
    \caption{Distribution of prediction outcomes across the evaluated image transformations. Samples are divided into those answered incorrectly under all tested conditions (\emph{all scores $=0$}), those whose predictions vary and are incorrect for the original image (\emph{mixed, originally wrong}), those whose predictions vary and are correct for the original image (\emph{mixed, originally correct}), and those answered correctly under all tested conditions (\emph{all scores $=1$}). A large proportion of benchmark errors are therefore recoverable under an alternative, semantically equivalent image representation.}
    \label{fig:score_distribution}
\end{figure}

Figure~\ref{fig:score_distribution} demonstrates that benchmark errors are not necessarily persistent across image transformations. For Llama-3.2-11B-Vision, for example, approximately half of the original predictions are incorrect. However, only $11.4\%$ of all samples remain incorrect under every tested condition, whereas $38.7\%$ are answered correctly under at least one alternative transformation. Thus, most of this model's original errors are recoverable under the tested variations and cannot be attributed solely to an invariant inability to solve the corresponding questions. Conversely, many originally correct predictions become incorrect after a change in image statistics, revealing a substantial source of instability that single-resolution accuracy does not capture.

This analysis demonstrates that two models with similar VQA accuracy may differ substantially in both the persistence of their errors and the stability of their correct predictions. Accuracy alone therefore provides an incomplete account of model behavior under resolution changes. These observations motivate AUSC and PVS as complementary metrics: AUSC measures the preservation of initially correct predictions, whereas PVS characterizes variation among initially incorrect predictions. In the following section, we compare these metrics with existing robustness measures and show that they capture distinct properties of model behavior.

\subsection{Comparison with Related Robustness Metrics}
\label{sec:metric_comparison}

We compare our metrics with the most closely related recent proposals:
ACE and RCE~\cite{li2026res} for AUSC, and
Flip$^{+}$ and Flip$^{-}$~\cite{saxena2026vlm} for PVS.
The differences are not merely definitional: as shown in
Table~\ref{tab:metric_comparison}, the metrics can produce different model
rankings because they capture distinct aspects of robustness to resolution
changes.

\begin{table*}[t]
    \centering
    \caption{
        Comparison of AUSC and PVS with existing robustness metrics.
        Higher values indicate better robustness for AUSC, whereas lower
        values indicate greater robustness for ACE, RCE, Flip$^{+}$, and
        Flip$^{-}$. For PVS, higher values indicate greater prediction
        variability among samples that are initially answered incorrectly.
    }
    \label{tab:metric_comparison}
    \resizebox{\textwidth}{!}{%
    \begin{tabular}{lrrrrrrr}
        \toprule
        Model
        & $\mathrm{Acc}_{\mathrm{orig}}$
        & AUSC
        & PVS
        & ACE
        & RCE
        & Flip$^{+}$
        & Flip$^{-}$ \\
        \midrule
        Qwen2.5-VL-7B-Instruct       & 0.604 & 0.863 & 0.368 & 0.054 & 0.091 & 0.086 & 0.067 \\
        Qwen2.5-VL-3B-Instruct       & 0.536 & 0.829 & 0.351 & 0.070 & 0.137 & 0.096 & 0.073 \\
        Qwen3-VL-4B-Instruct         & 0.600 & 0.827 & 0.457 & 0.086 & 0.151 & 0.109 & 0.079 \\
        Qwen3-VL-8B-Instruct         & 0.627 & 0.816 & 0.408 & 0.109 & 0.195 & 0.121 & 0.063 \\
        Qwen3-VL-2B-Instruct         & 0.466 & 0.816 & 0.409 & 0.080 & 0.174 & 0.090 & 0.085 \\
        Qwen2-VL-7B-Instruct         & 0.562 & 0.810 & 0.363 & 0.109 & 0.214 & 0.112 & 0.070 \\
        Qwen2-VL-2B-Instruct         & 0.448 & 0.793 & 0.326 & 0.078 & 0.191 & 0.097 & 0.072 \\
        Llama-3.2-11B-Vision-Instruct
                                     & 0.499 & 0.669 & 0.477 & 0.173 & 0.429 & 0.173 & 0.083 \\
        LLaVA-1.5-7B-HF              & 0.352 & 0.666 & 0.326 & 0.102 & 0.348 & 0.123 & 0.060 \\
        Pixtral-12B                  & 0.499 & 0.664 & 0.427 & 0.170 & 0.435 & 0.171 & 0.083 \\
        \bottomrule
    \end{tabular}%
    }
\end{table*}

\paragraph{AUSC versus ACE and RCE.}
ACE and RCE are computed over the complete evaluation set and quantify accuracy changes between adjacent resolution levels. Consequently, losses on samples that are initially answered correctly can be offset by gains on initially incorrect samples. AUSC instead conditions on initially correct samples and measures the retention of these predictions over the complete resolution range. Thus, ACE and RCE characterize local changes in the accuracy curve, whereas AUSC captures overall preservation of initially correct predictions.

This distinction can change the resulting model ranking. For example, ACE ranks Qwen2-VL-2B as more robust than Qwen3-VL-8B ($0.078$ versus $0.109$), whereas AUSC reverses this ordering ($0.793$ versus $0.816$). Qwen3-VL-8B exhibits larger changes between adjacent resolutions but retains a greater proportion of its initially correct predictions over the full resolution range. Metrics that emphasize local variation therefore do not necessarily capture the global performance retention measured by AUSC.

\paragraph{PVS versus Flip rates.}
The principal difference between PVS and Flip$^{+}$/Flip$^{-}$ is the normalization population. Flip rates are computed relative to the complete dataset and therefore depend on baseline accuracy: a model with lower initial accuracy has a larger pool of initially incorrect samples that may change their predictions. PVS instead normalizes by the number of initially incorrect samples, isolating their conditional prediction variability and making comparisons across models with different baseline accuracies more meaningful.

Table~\ref{tab:metric_comparison} illustrates the effect of this normalization. Flip$^{-}$ identifies Qwen3-VL-2B as slightly more unstable than Llama-3.2-11B ($0.085$ versus $0.083$), whereas PVS reverses the ordering ($0.409$ versus $0.477$). Normalizing Flip$^{-}$ by the fraction of initially incorrect samples,
\[
    \frac{\mathrm{Flip}^{-}}
         {1-\mathrm{Acc}_{\mathrm{orig}}},
\]
produces the same reversal: the normalized values are $0.159$ for Qwen3-VL-2B and $0.166$ for Llama-3.2-11B. Hence, the higher unnormalized Flip$^{-}$ value of Qwen3-VL-2B primarily reflects its lower baseline accuracy rather than greater conditional variability among initially incorrect samples.

Overall, these comparisons show that AUSC and PVS are not interchangeable with existing metrics. AUSC separates the preservation of initially correct predictions from compensating accuracy gains, while PVS separates conditional prediction variability from the prevalence of baseline errors. These properties are central to attributing resolution-induced failures to the behavior of a model rather than to differences in its initial accuracy.

\subsection{Per-Benchmark Analysis of Resolution Sensitivity}
\label{sec:per_benchmark_analysis}

Aggregate AUSC and PVS scores summarize model behavior across benchmarks but can conceal substantial variation between visual tasks. We therefore disaggregate both metrics by benchmark and transformation type. This analysis examines whether prediction stability depends on benchmark characteristics and whether different image transformations expose distinct failure modes.

\begin{figure}
    \centering
    \includegraphics[width=0.9\linewidth]{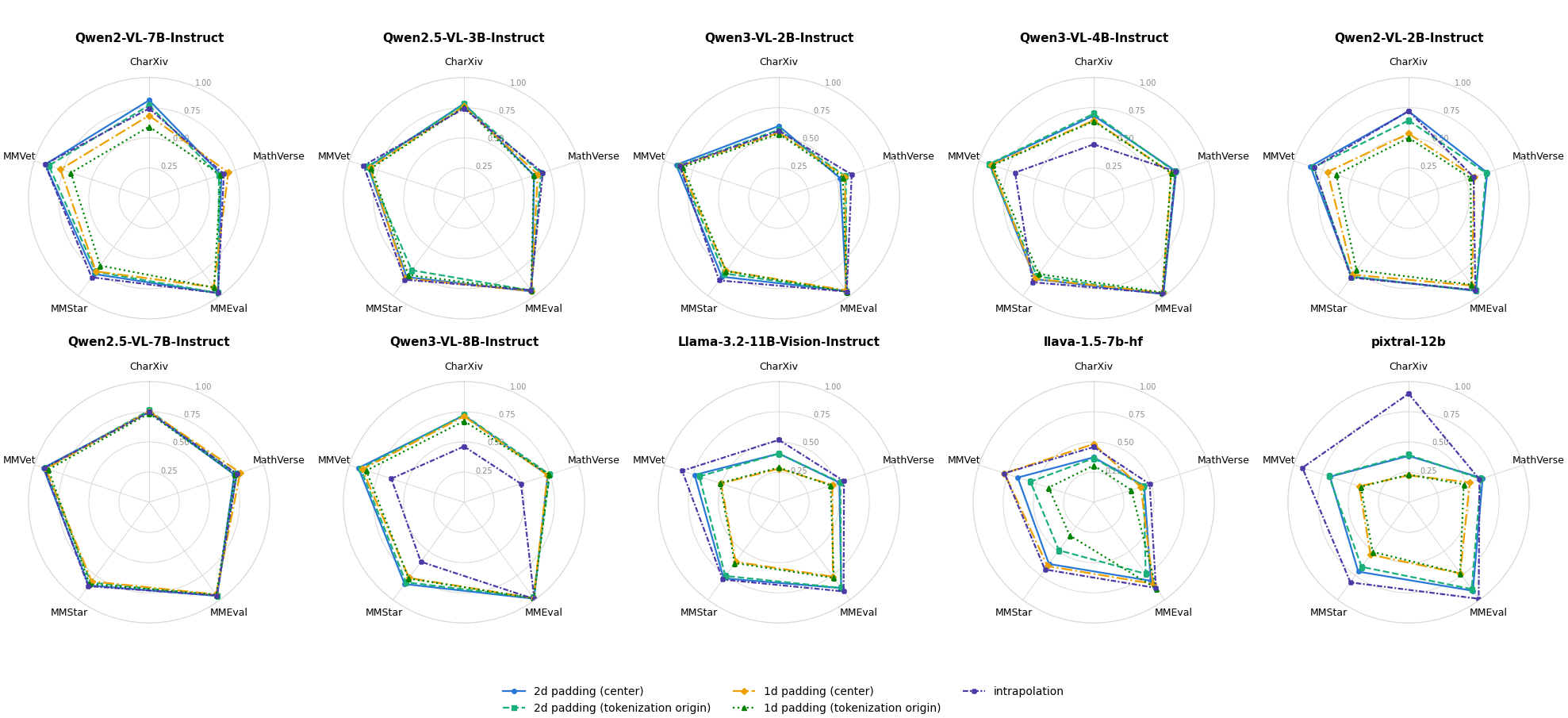}
    \caption{AUSC disaggregated by benchmark and transformation type. Each panel corresponds to one model, and each axis to one benchmark. Higher values indicate stronger retention of predictions that are correct on the original images. The five curves represent centered and tokenization-aligned 1D and 2D padding, as well as interpolation.}
    \label{fig:ausc_per_benchmark}
\end{figure}

\begin{figure}
    \centering
    \includegraphics[width=0.9\linewidth]{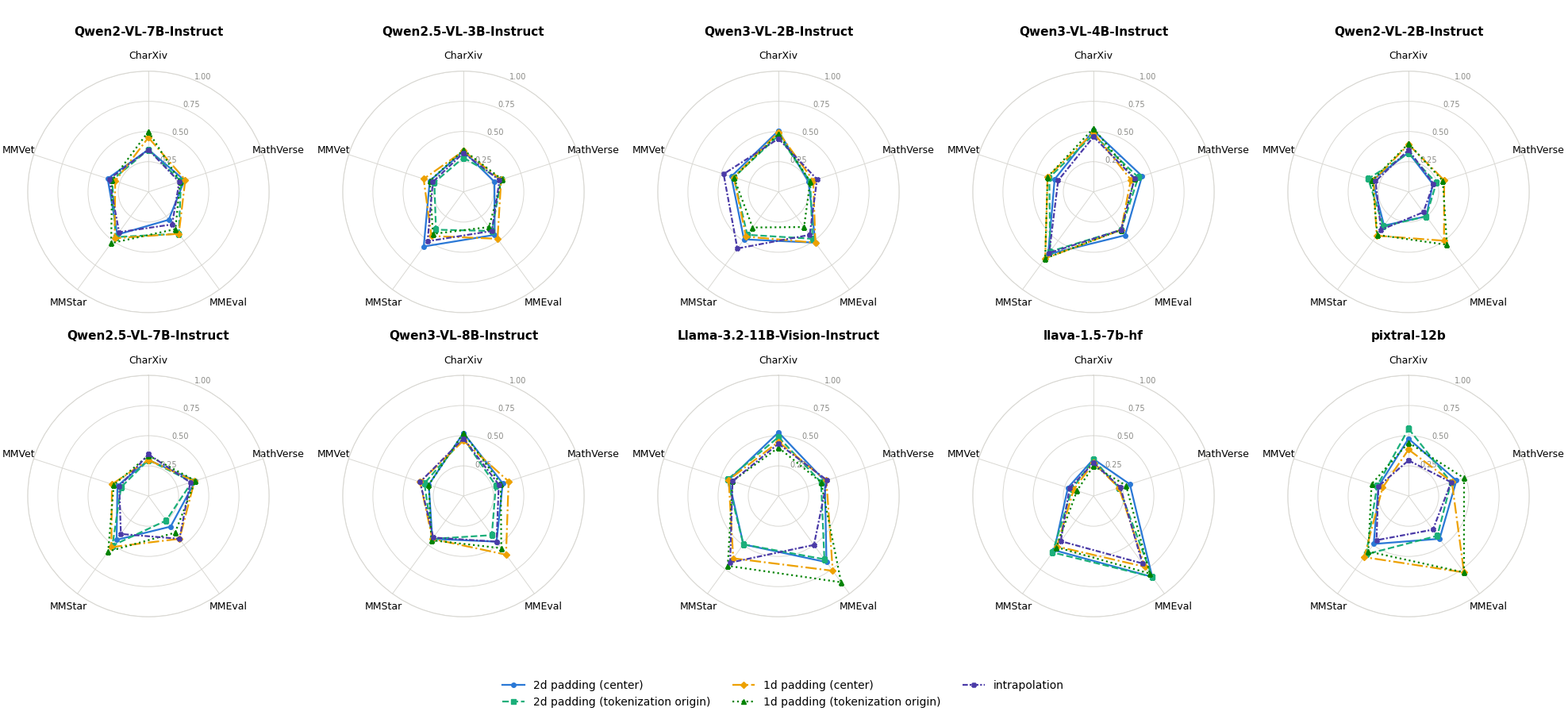}
    \caption{PVS disaggregated by benchmark and transformation type. Higher values indicate greater prediction variability among samples that are answered incorrectly on the original images. Differences between transformation types reveal that the recoverability of these errors depends on how the image representation is modified.}
    \label{fig:pvs_per_benchmark}
\end{figure}

\paragraph{Stability depends on benchmark characteristics.}
Figure~\ref{fig:ausc_per_benchmark} reveals pronounced benchmark-specific differences in the preservation of initially correct predictions. Across many of the Qwen models, AUSC is highest on MMEval, MMStar, and MMVet, while greater degradation is frequently observed on CharXiv and MathVerse. However, the magnitude of these differences remains model dependent; no benchmark is uniformly more or less stable across all architectures.

These results are consistent with the expected role of task-relevant detail size. Predictions should remain stable for questions determined by large visual structures or the overall scene because the relevant evidence remains accessible across a wider range of transformations. In contrast, questions whose answers depend on a small localized region are more sensitive to changes in resolution, interpolation, and patch alignment. Benchmark composition can therefore affect measured robustness even when the transformations preserve the semantic content of the image. Because the benchmarks are not explicitly annotated by the spatial extent of their task-relevant evidence, this interpretation should be viewed as a plausible explanation of the observed trend rather than a direct causal attribution.

\paragraph{Transformation discrepancies reveal model-specific behavior.}
The distance between transformation-specific curves provides further information about potential failure mechanisms. For Qwen2.5-VL-3B and Qwen2.5-VL-7B, the AUSC curves largely overlap, suggesting that degradation is relatively insensitive to the particular transformation used. Other models exhibit substantially larger discrepancies. Most notably, interpolation produces considerably lower AUSC than padding for Qwen3-VL-8B on several benchmarks, whereas the opposite pattern appears for Pixtral-12B, for which interpolation is often substantially less damaging. LLaVA-1.5 also exhibits large differences between centered and tokenization-aligned padding on several benchmarks.

The absence of a universally most damaging transformation indicates that these effects cannot be explained by input size alone. Instead, they likely reflect interactions among the transformation, the model's image preprocessing pipeline, its tokenization or patchification strategy, and the visual encoder architecture. Consequently, evaluating only one type of resolution-preserving transformation may overlook model-specific vulnerabilities.

\paragraph{PVS exposes asymmetric behavior on initially incorrect samples.} The per-benchmark PVS results in Figure~\ref{fig:pvs_per_benchmark} do not simply mirror the AUSC patterns. A model may preserve its initially correct predictions while simultaneously exhibiting considerable variability among its initially incorrect predictions. This behavior is especially visible on MMEval: several models achieve high AUSC on this benchmark but also show high
PVS under particular transformations. Thus, robustness of correct predictions and recoverability of incorrect predictions represent distinct properties and should be analyzed separately.

PVS also exhibits pronounced transformation-specific asymmetries. For example, the variability of Llama-3.2-11B-Vision on MMEval is substantially higher under tokenization-aligned 1D padding than under interpolation. Pixtral-12B shows a similarly large separation on MMEval, with 1D padding
producing markedly higher PVS than interpolation. In contrast, several Qwen models display comparatively aligned PVS curves, indicating less dependence on the precise transformation. These asymmetries suggest that an image
modification can recover initially incorrect predictions for one architecture while having little effect for another.

Overall, the per-benchmark results show that resolution sensitivity is jointly determined by the benchmark, model, and image transformation. Aggregate metrics remain useful for ranking models, but disaggregated AUSC and PVS expose where instability occurs and whether it primarily affects initially correct or initially incorrect predictions.

\subsection{Per Image Size Analysis}

\begin{figure}
    \centering
    \includegraphics[width=0.9\linewidth]{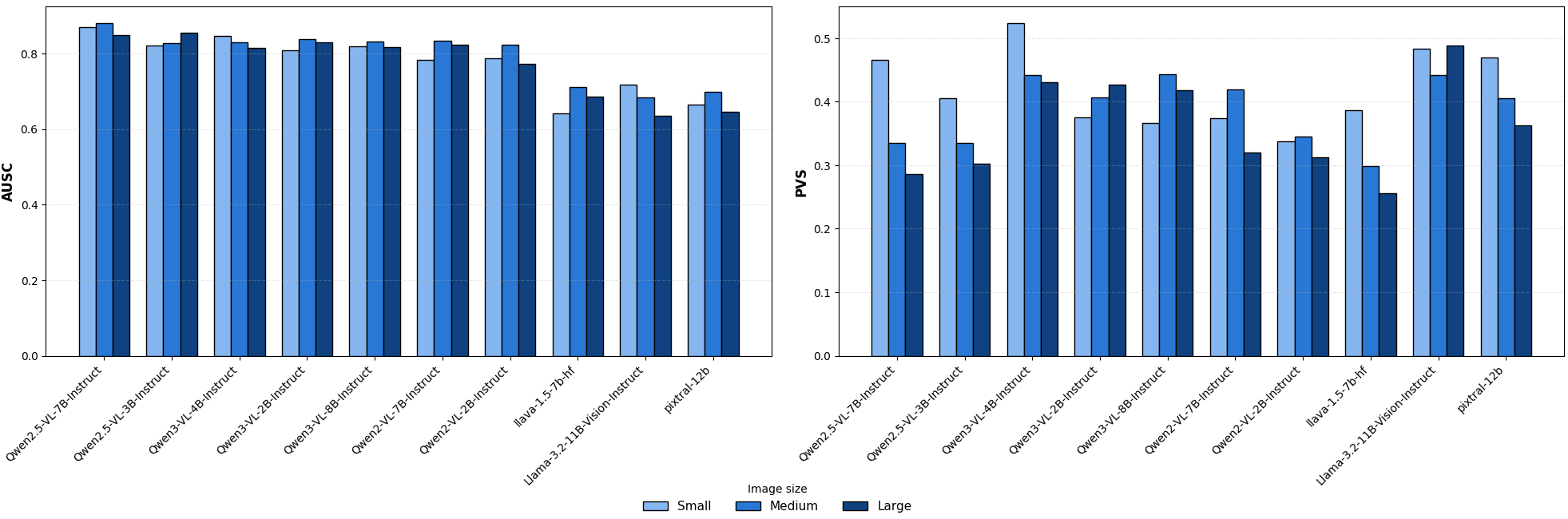}
    \caption{AUSC and PVS stratified by original image size. AUSC varies only modestly across size strata, whereas PVS generally decreases from small to large images, indicating stronger transformation sensitivity among initially incorrect predictions on smaller inputs. This trend is model dependent and is not observed uniformly across architectures.}
    \label{fig:metrics_by_image_size}
\end{figure}

Figure~\ref{fig:metrics_by_image_size} shows that AUSC is comparatively stable across image-size strata, with no consistent monotonic relationship between original image size and the retention of initially correct predictions. PVS exhibits a stronger size dependence: for seven of the ten models, PVS is lower for large than for small images, with particularly pronounced decreases for Qwen2.5-VL-7B, LLaVA-1.5, and Pixtral-12B. This suggests that initially incorrect predictions on smaller images are more likely to change under the evaluated transformations. The exceptions, including Qwen3-VL-2B and Llama-3.2-11B-Vision, show that this effect remains architecture dependent. Overall, original image size explains more variation in the behavior of initially incorrect predictions than in the preservation of initially correct ones.

\section{Human Review of Augmented Images}
To verify that the evaluated transformations preserve semantic content, we conducted a human validation study on a subset of transformed images. We randomly sampled images from all benchmarks and applied the same padding and bicubic interpolation procedures used in the main experiments. Human annotators were shown the original and transformed images side-by-side and asked whether the transformation altered the semantic interpretation required to answer the task correctly.

Across all evaluated samples, annotators judged the vast majority of padded and interpolated images as semantically equivalent to the originals. Padding transformations were consistently rated as preserving semantic content, while bicubic interpolation occasionally introduced mild blur but rarely changed task-relevant information. In contrast, preliminary experiments with super-resolution methods produced noticeable hallucinations and texture artifacts, particularly in fine-grained regions, and were therefore excluded from the final evaluation.

These results support our assumption that the transformations used in the paper primarily modify resolution characteristics and information density while preserving the underlying semantic content of the image.

\section{Detailed Metric Specification}\label{appendix:metrics}
\subsection{Area Under the Scaling Curve (AUSC)}

\begin{figure}[h]
\centering
\includegraphics[width=0.6\linewidth]{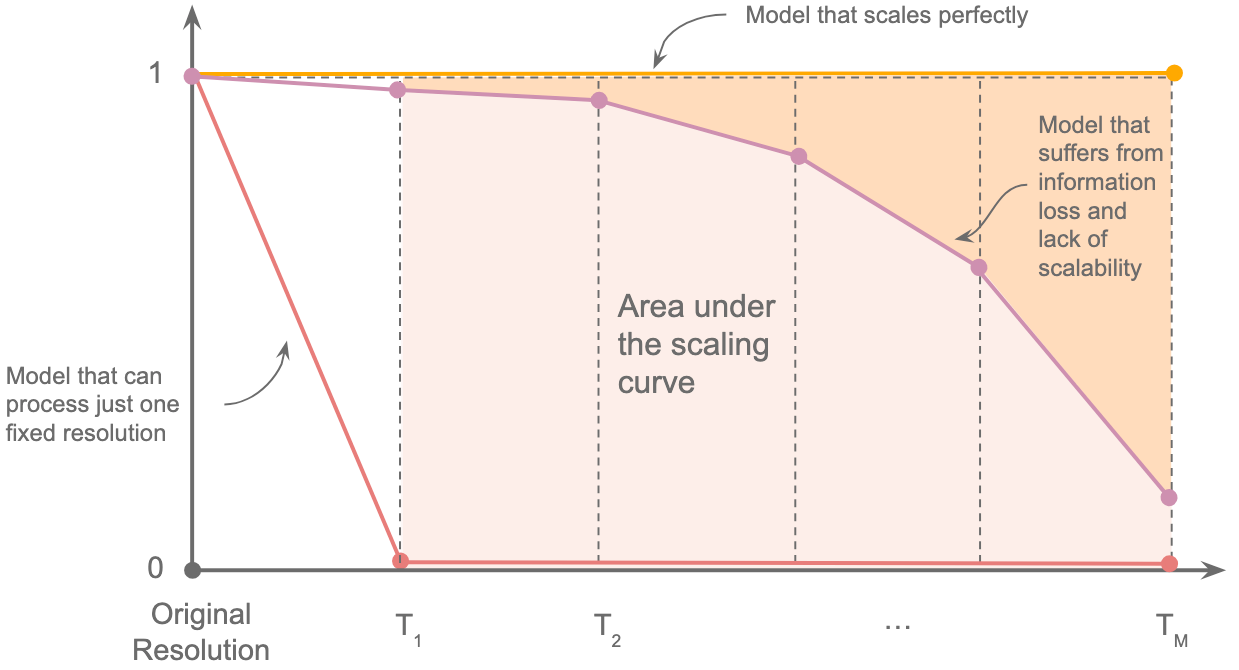}
\caption{Illustration of the Area Under the Scaling Curve (AUSC) metric. AUSC is computed on $\mathcal{B}_{\text{correct}}$ and measures the area under the accuracy-vs-scale curve, normalized to $[0,1]$.}
\label{fig:ausc}
\end{figure}

Let \( \mathcal{B}_{\text{correct}} \subseteq \mathcal{B} \) denote the subset of image-question pairs for which the model provides the correct answer at the original resolution:

\begin{equation}
\mathcal{B}_{\text{correct}} = \{(x_i, y_i) \in \mathcal{B} \mid f(x_i, y_i) = z_i \}
\label{eq:correct_batch:appendix}
\end{equation}

Let \( \mathcal{T} = \{T_{1}, T_{2}, \dots, T_M\} \) be a set of image transformations as described in Section 3.2, where \(M\) represents the number of transformation levels (e.g., \(M = 8\) for padding with margins from \(0.25×\) to \(2.0×\) image size in \(0.25×\) increments). For each transformation \(T_j\), we construct a transformed benchmark subset:

\begin{equation}
\mathcal{B}_{\text{correct}}^{(j)} = \left\{ \left( T_j(x_i), y_i \right) \mid (x_i, y_i) \in \mathcal{B}_{\text{correct}} \right\}
\label{eq:correct_batch_transformation}
\end{equation}

For each size of the margin \( j = 1, 2, ..., M \), we evaluate the model \( f(x) \) based on the chosen metric and evaluation protocol (for example, VQA accuracy). 

Given the sequence of accuracies \(\text{Acc}(T_1), \text{Acc}(T_2), \ldots, \text{Acc}(T_M)\}\), we calculate the area under the scaling curve using the trapezoidal rule:

\begin{equation}
\text{AUSC} = \frac{1}{M - 1} \sum_{j=1}^{M - 1} \frac{1}{2}\biggl[\text{Acc}(T_{j}) + \text{Acc}(T_{j+1})\biggr]
\label{eq:AUSC:appendix}
\end{equation}

\textbf{Interpretation. }AUSC values lie in \( \in [0, 1] \) and have an intuitive interpretation:

\begin{itemize}
    \item \textbf{AUSC} \( = 1 \): The model scales perfectly under the set of transformations $\mathcal{T}$. This indicates \emph{maximum robustness}.
    
    \item \textbf{AUSC} \( \in (0, 1) \): Model architecture and image preprocessing steps incur visual information loss. Higher AUSC values reflect smoother degradation and more stable performance across transformations. Lower AUSC values indicate that the model is sensitive to certain transformations and quickly degrades.
    
    \item \textbf{AUSC} \( = 0 \): The model can process just one fixed resolution and fails to recognize any information in images of higher resolutions. This indicates \emph{complete brittleness}.
\end{itemize}

\textbf{Statistical Significance.} One of the limitations of the AUSC is its dependence on the size of \(\mathcal{B}_{\text{correct}}\). For the purpose of statistical significance, we recommend
\(|\mathcal{B}_{\text{correct}}| \geq 75\).

\subsubsection{Prediction Variance Score (PVS)}

For additional insight, we assess how variable the predictions are under the set of transformations \(\mathcal{T}\) when considering instances that the model originally did not answer correctly. This reveals whether resolution changes can "rescue" incorrect predictions, indicating sensitivity to tokenization or visual encoding artifacts.

Let \(\mathcal{B}_{\text{wrong}} = \mathcal{B} \setminus \mathcal{B}_{\text{correct}}\) denote examples where the model initially predicted incorrectly. For each transformation \(T_j\), we calculate the Resolution Bias Score as:

\begin{equation}
\text{PVS}(T_j) = \frac{1}{|\mathcal{B}_{\text{wrong}}|} \sum_{(x_i, y_i) \in \mathcal{B}_{\text{wrong}}} \mathds{1}_{\text{correct}}\bigl[f(T_j(x_i), y_i) = z_i\bigr]
\label{eq:PVS-details}
\end{equation}

\textbf{Interpretation.} This metric captures how often the transformation \(T_j\) causes the model to produce the correct answer on an image-question pair for which the model produced an incorrect prediction at the original resolution. The PVS lies in \( [0, 1] \):

\begin{itemize}
\item \textbf{PVS} \( = 0 \): The model is invariant toward shifts and size changes of visual information. Incorrect predictions remain incorrect regardless of resolution.
\item \textbf{PVS} \( \in (0, 1) \): Resolution changes flip predictions from incorrect to correct. For example, \(\text{PVS}(T_j) = 0.3\) means that 30\% of originally incorrect answers became correct after applying transformation \(T_j\).
\item \textbf{PVS} \( = 1 \): All incorrect predictions become correct after transformation (extremely rare and would indicate severe resolution bias).
\end{itemize}

\textbf{Diagnostic Value.} High PVS values may signal:

\begin{itemize}
\item \textbf{Potential data augmentation benefits:} The vision encoder could benefit from training with similar transformations
\item \textbf{Lack of robustness:} Shifts in model attention or reasoning pathways, indicating brittleness toward image transformations
\item \textbf{Tokenization artifacts:} Token boundary alignment critically affects model predictions
\end{itemize}

Similarly to AUSC, we recommend to choose sufficiently large and challenging benchmarks with \(|\mathcal{B}_{\text{wrong}}| \geq 75\).

\section{LLM-as-a-Judge Evaluation}
\label{app:llm_judge}

To evaluate semantic equivalence between model predictions and ground-truth answers across heterogeneous benchmarks, we employ an LLM-as-a-judge protocol using \texttt{GPT-5-nano}. The judge model receives the question, the ground-truth answer and the model prediction and outputs a binary correctness score.

The following prompt was used for all evaluations:

\begin{quote}
\small
\begin{verbatim}
Compare two answers to the question and output 1 if the
predicted answer contains the same information as the
ground truth. Output 0 if some of the information is
missing or the answer is completely wrong.

ex1:
Ground truth: "a"
Prediction: "the style depicted in this image is a:
impressionism."
Score: 1

ex2:
Ground truth: "$\frac(1)(2000)$"
Prediction: "the average rate of change in the graph
is 5."
Score: 0

Is the predicted answer correct?

Question: {question}
Ground truth: {ground_truth}
Prediction: {prediction}

Output only the numeric score (0 or 1)
\end{verbatim}
\end{quote}

The binary outputs produced by the judge model were used as the final correctness labels in all experiments requiring open-form answer matching.

To validate the reliability of the automatic evaluation procedure, we manually reviewed 500 randomly sampled answer pairs spanning all benchmarks and transformation settings. Human annotations were compared against the \texttt{GPT-5-nano} judgments, yielding an agreement accuracy of over \textbf{97\%}. These results indicate that the automated evaluation protocol provides a reliable approximation of human judgment.

We must note that this type of evaluation is sensitive to the used prompt and the choice of the judge model. 

\section{Compute Resources}

All experiments and evaluations were conducted using a compute server equipped with 4$\times$NVIDIA RTX A6000 GPUs. The evaluation pipeline is also compatible with a single-GPU setup, although runtimes increase proportionally due to the large number of evaluated transformations and model inferences.

%%%%%%%%%%%%%%%%%%%%%%%%%%%%%%%%%%%%%%%%%%%%%%%%%%%%%%%%%%%%

\end{document}